\documentclass[sigconf]{acmart}
\AtBeginDocument{%
  \providecommand\BibTeX{{%
    \normalfont B\kern-0.5em{\scshape i\kern-0.25em b}\kern-0.8em\TeX}}}

\copyrightyear{2023} 
\acmYear{2023} 
\setcopyright{acmlicensed}\acmConference[ICAIL 2023]{Nineteenth International Conference on Artificial Intelligence and Law}{June 19--23, 2023}{Braga, Portugal}
\acmBooktitle{Nineteenth International Conference on Artificial Intelligence and Law (ICAIL 2023), June 19--23, 2023, Braga, Portugal}
\acmPrice{15.00}
\acmDOI{10.1145/3594536.3595163}
\acmISBN{979-8-4007-0197-9/23/06}

\usepackage{wrapfig}
\usepackage{xcolor}
\usepackage{soul}
\newcommand{\hlc}[2][yellow]{{%
    \colorlet{foo}{#1}%
    \sethlcolor{foo}\hl{#2}}%
}
 \usepackage{multirow} % More at https://texblog.org/2012/12/21/multi-column-and-multi-row-cells-in-latex-tables/
 \usepackage{setspace} % for \onehalfspacing and \singlespacing macros

\begin{document}

%\title{Can GPT-3 Do Statutory Reasoning?}

\title{Can GPT-3 Perform Statutory Reasoning?}

\author{Andrew Blair-Stanek}
\email{ablair-stanek@law.umaryland.edu}
\affiliation{%
  \institution{University of Maryland School of Law}
  \city{Baltimore}
  \state{Maryland}
  \country{USA}
}

\author{Nils Holzenberger}
\email{nils.holzenberger@telecom-paris.fr}
\affiliation{%
  \institution{T\'el\'ecom Paris\\ Institut Polytechnique de Paris}
  \city{Palaiseau}
  \country{France}
}

\author{Benjamin Van Durme}
\email{vandurme@jhu.edu}
\affiliation{%
  \institution{Johns Hopkins University}
  \city{Baltimore}
  \state{Maryland}
  \country{USA}
}

\renewcommand{\shortauthors}{Blair-Stanek et al.}

\begin{abstract}
   Statutory reasoning is the task of reasoning with facts and statutes, which are rules written in natural language by a legislature.  It is a basic legal skill.  In this paper we explore the capabilities of the most capable GPT-3 model, \texttt{text-davinci-003}, on an established statutory-reasoning dataset called SARA.  We consider a variety of approaches, including dynamic few-shot prompting, chain-of-thought prompting, and zero-shot prompting.   While we achieve results with GPT-3 that are better than the previous best published results, we also identify several types of clear errors it makes.  We investigate why these errors happen.  We discover that GPT-3 has imperfect prior knowledge of the actual U.S. statutes on which SARA is based.  More importantly, we create simple synthetic statutes, which GPT-3 is guaranteed not to have seen during training.  We find GPT-3 performs poorly at answering straightforward questions about these simple synthetic statutes. % We find that even with 2-shot reasoning prompts to guide it, GPT-3 struggles with all but the most basic synthetic statutes. 
   
\end{abstract}




%\received{12 February 2023}

\maketitle

\definecolor{GPThighlight}{rgb}{0.83, 0.95, 0.83} % This is the color GPT-3's playground uses to highlight its own responses.  

\section{Introduction}

\begin{figure}
    \hrule
    \flushleft
        \vspace{1ex}

Section 1001.  Definition of Rolang.

    (a) General rule

The term "rolang" means-

    \hspace{4ex} (1) any soultratessly, or

    \hspace{4ex} (2) any parkinse.

(b) Soultratessly

    The term "soultratessly" means-

    \hspace{4ex} (1) any oxideney, or

    \hspace{4ex} (2) any chastiles.

    (c) Parkinse

    The term "parkinse" means-

    \hspace{4ex} (1) any portle, or

    \hspace{4ex} (2) any frestes.

        \vspace{1ex}

    Alexis is a portle. Is section 1001(b) applicable to Alexis? Let's think step by step.

        \vspace{1ex}

    \hlc[GPThighlight]{Section 1001(a) defines rolang as either soultratessly or parkinse.}
    
        \vspace{1ex}
    
    \hlc[GPThighlight]{Section 1001(b) defines soultratessly as either oxideney or chastiles.}

        \vspace{1ex}

    \hlc[GPThighlight]{Section 1001(c) defines parkinse as either portle or frestes.}

        \vspace{1ex}

    \hlc[GPThighlight]{Since Alexis is a portle, section 1001(b) is applicable to Alexis.}
            \vspace{1ex}

    \hrule

\caption{Example of GPT-3 incorrectly answering a simple question about a short synthetic statute. The text generated by GPT-3 is \hlc[GPThighlight]{highlighted}. }
% We refer the task details to \cref{sec::tasks}.}

    \label{fig:simple_gpt3_fail}

\end{figure}

There has been great excitement about the emergent capabilities of large language models such as GPT-3 \citep{brown20language} and  ChatGPT on legal tasks.  For example, \citet{bommarito22gpt} show that GPT-3 performs at a passing rate on evidence and torts questions on the U.S. multistate bar exam.  In \citet{choiGPTfinal},  law professors had ChatGPT take their final exams and blindly graded its output along with their human students' answers; it achieved low but passing grades.  

We set out to test how the most powerful GPT-3 model currently exposed through public APIs would perform on one of the most basic tasks required of lawyers, \emph{statutory reasoning}.  Statutes are legal rules written by legislative bodies in natural language.  Statutory reasoning is the application of those rules to facts that are also in natural language~\citep{lawsky17logic}.
We first consider the StAtutory Reasoning Assessment (SARA) dataset, a handcrafted computational benchmark covering nine sections of the U.S. tax code~\citep{holzenberger20dataset}.   We find that straightforward application of GPT-3 leads to significant gains compared to prior work on this task, but that GPT-3 makes a number of clear mistakes (\S\ref{sec:sara}).  

We then ask why this could be, 
%set out to understand the explanations for our results.
starting with the question of how much GPT-3 already knows about the current U.S. Code (\S\ref{sec:knowsIRC}).  We find that it has some imperfect knowledge of the U.S. Code. %that outdated versions of the code were part of GPT-3's training data. 
Then we detail (\S\ref{sec:synthetic})  an experiment based on novel synthetic statutes to measure how well GPT-3 can answer simple statutory-reasoning questions when prompted with a single fact (see Figure~\ref{fig:simple_gpt3_fail} for an example). When presented with statutory language that is easily interpretable by humans, we find that GPT-3 will regularly make mistakes. Our results suggest statutory reasoning as an area of interest for new AI research, as a challenge to motivate future improvements in large language models.

\section{Related work}

\paragraph{Statutory reasoning}

The SARA dataset~\citep{holzenberger20dataset} is a benchmark for statutory reasoning%, and to our knowledge the only such benchmark 
. Efforts to solve SARA through machine reading methods have had limited success~\citep{holzenberger20factoring}. SARA's curated statutes, facts, and questions were designed to avoid semantic ambiguity.  But semantic ambiguity also poses challenges for statutory reasoning, such as the U.S. Supreme Court case \textit{Nix v. Hedden} about whether tomatoes were "vegetables" in a customs statute. There have been several approaches for using NLP to handle statutory reasoning involving such semantic ambiguity \citep{vsavelka2021legal,vsavelka2021discovering}.  The task of determining whether one statute entails another one, called \emph{statute entailment}, is the closest legal NLP task to statutory reasoning. The yearly COLIEE challenge~\citep{rabelo21overview} includes statute entailment. BERT-based models~\citep{devlin19bert} have been most successful on COLIEE's statute entailment, until they were recently surpassed by GPT-3-based models~\citep{yu22legal}.
 Legal judgment prediction is another task related to statutory reasoning, and is also a significant challenge for current machine reading models~\citep{zhong20iteratively,lam20gap,bi22judicial}.

\paragraph{Prompting GPT-3}

GPT-3 is a large-scale language model, which has shown a surprising ability to solve NLP tasks without the need to fine-tune its parameters~\citep{brown20language}.
This is achieved by providing GPT-3 with a set of natural language instructions, specifying what task is to be solved, along with a few input-output examples for the task.
This prompt-based approach has been further refined, in particular by augmenting the expected output with human-written natural language explanations of how that output was derived from the input~\citep{wei22chain}.
\citet{kojima22large} show that GPT-3 can be prompted to produce these explanations itself, providing a substitute for human annotation.
In particular, the phrase "Let's think step by step" was found, out of a number of prompts, to maximize performance.
We prompt GPT-3 in multiple ways, taking inspiration from \citet{wei22chain} and \citet{kojima22large}.

\paragraph{Reasoning with GPT-3}
Exploring GPT-3's ability to reason has yielded improvements on multiple benchmark datasets.
\citet{lu22learn} collect a dataset of common sense and science questions, pairing each question with background information and an explanation of the answer. The additional context, serving as a chain-of-thought, marginally improves GPT-3's performance.
\citet{zhou22least} explicitly decompose reasoning problems, forcing GPT-3 to solve a complex problem in multiple increments.
\citet{khot22decomposed} show GPT-3 can be prompted to decompose problems into simpler steps, on both artificial tasks and open-domain question-answering.
\citet{zelikman22star} exploit a language model's ability to generate chains of reasoning: GPT-J~\citep{wang21gpt} is made to generate chains of reasoning for a set of questions and answers, and is further fine-tuned on those chains that lead to the correct answer.
Adding structure to the process of natural language generation, \citet{zhang22improved} run a symbolic reasoning engine on the output of a language model, yielding improvements on semi-synthetic benchmarks meant to test reasoning abilities.
\citet{thorburn22optimizing} evaluate GPT Neo's argumentative abilities, by making it suggest claims or reasons to support a claim, with no particular emphasis on logical reasoning.
Data comes from Kialo, a collection of arguments curated by online users.
Evaluations show that reasons and claims generated by GPT Neo are less coherent than those generated by humans.
\citet{saparov22language} systematically study GPT-3's ability to reason, using synthetic, controlled data, and metrics that measure the internal coherence of chains of reasoning. The authors show that GPT-3 has the ability to perform individual steps of reasoning, but tends to lose track when producing a proof with multiple steps. Results further show that GPT-3 leverages its background knowledge about the world to draw conclusions, so that GPT-3's ability to reason decreases as the overlap between the target domain and its real-world knowledge decreases.

\paragraph{GPT-3 on legal tasks}
\citet{hendrycks21measuring} use GPT-3 to solve legal multiple-choice questions, with data collected from online sources. Results show that the largest GPT-3 model performs significantly better than random, but still far less than expert human performance.
\citet{yu22legal} experiment with various ways of prompting GPT-3 for statute entailment.
The authors find that \mbox{GPT-3} does much better than previous BERT-based models.
In particular, prompts designed with inspiration from legal reasoning work best.
\citet{bommarito22gpt} test GPT-3's ability to answer multiple choice questions, involving short paragraphs of context, from the U.S. multistate bar exam. Prompting GPT-3 to rank the available answer choices gets close to a passing score on some subject areas.   In the medical domain, \citet{gutierrez22thinking} study GPT-3's abilities on two biomedical information extraction tasks. The authors show that fine-tuning a BERT model on a small training set consistently outperforms few-shot learning with GPT-3.

\section{SARA}
\label{sec:sara}

\begin{figure*}
\includegraphics[width=\textwidth]{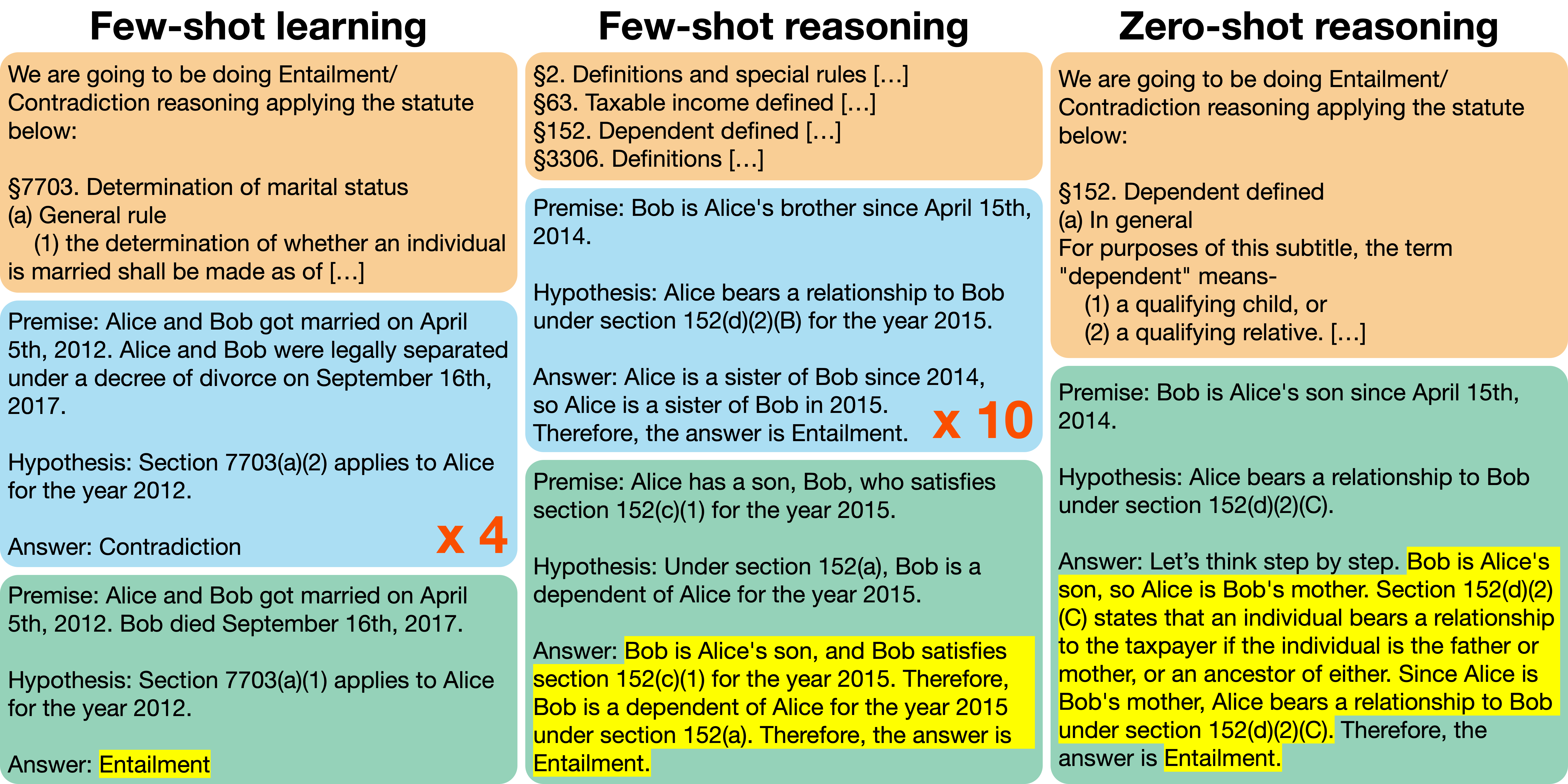} 
\caption{Prompts used in Section~\ref{sec:sara} to pose SARA test cases to GPT-3. Top boxes, in orange, contain statutes; depending on the setting, statutes may or may not be included. Example cases are in blue; in zero-shot there are no example cases. At the bottom, in green, are test cases. Text highlighted in yellow is generated by GPT-3. If GPT-3's first response is not clear, we prompt a second time with "Therefore the answer is", following \citet{kojima22large}. }
\label{fig:prompting-GPT-3-SARA}
\end{figure*}

\subsection{Data and Experimental Setup}

SARA stands for StAtutory Reasoning Assessment.  It consists of nine statutory sections drawn from the U.S. tax code (i.e. title 26 of the U.S. Code) and 376 hand-crafted cases that state simple facts, involving taxpayers named Alice, Bob, Charlie, and Dan, and ask one question that can be answered by applying some of the nine sections to the facts.  Obviously, the nine statutes, the facts, and the questions are all in natural language.  But SARA also includes a translation of the statutes, facts, and questions to Prolog. Solving the questions via Prolog results in 100\% accuracy.  

Of the 376 cases, 100 are pure "tax cases" and ask for how much tax an individual had to pay, with answers typically in the many thousands of dollars.  Because GPT-3 performs poorly in reasoning with such numbers, we do not consider these 100 cases.  All the remaining 276 cases ask for either "Entailment" or "Contradiction".  Of these, 193 (121 train, 72 test) involve numbers.  The remaining 83 (55 train, 28 test) involve no numbers at all.  For all our tests, we break out our results separately on the 72 test cases with numbers and the 28 test cases with no numbers.  

In all experiments throughout this paper, for our calls to GPT-3, we set the temperature to 0.0 and the top\_P to 1.0.\footnote{All our code is at github.com/BlairStanek/gpt-statutes.} These settings serve two purposes.  First, they maximize reproducibility.  Second, they minimize the risk that GPT-3 will wander off topic or hallucinate.
In SARA, every question is annotated with a single correct answer.

\begin{table*}
\begin{center}
\begin{tabular}{ r c c c c c } 
 \toprule
 & Include & "Let's think & \multicolumn{2}{c}{Accuracy in \% on cases with:} & Aggregate\\
Prompt type & Statute & step by step" &   numbers & no numbers & accuracy \\
 \toprule
\multirow{4}*{4-shot dynamic} & Y & Y  & 60 $\pm$ 10 (43/72) & 61 $\pm$ 16 (17/28) & 60 $\pm$ 8 (60/100) \\
 & Y &  N & 47 $\pm$ 10 (34/72) & 50 $\pm$ 16 (14/28) & 48 $\pm$ 8 (48/100) \\
 & N & Y  & 47 $\pm$ 10 (34/72) & 50 $\pm$ 16 (14/28) & 48 $\pm$ 8 (48/100) \\
 & N &  N & 49 $\pm$ 10 (35/72) & 64 $\pm$ 16 (18/28) & 53 $\pm$ 8 (53/100) \\
 \midrule
\multirow{4}*{zero-shot}  & Y & Y  & 61 $\pm$ 10 (44/72) & 75 $\pm$ 14 (21/28) & 65 $\pm$ 8 (65/100) \\
 & Y & N  &  74 $\pm$ \phantom{0}9 (53/72) & 64 $\pm$ 16 (18/28) & 71 $\pm$ 8 (71/100) \\
 & N & Y  & 49 $\pm$ 10 (35/72) &  50 $\pm$ 16 (14/28) &  49 $\pm$ 8 (49/100) \\
 & N & N  & 38 $\pm$ 10 (27/72) & 39 $\pm$ 16 (11/28) & 38 $\pm$ 8 (38/100) \\
 \midrule
\multirow{2}*{10-shot hand-crafted chain-of-thought}  & Y &  N & 56 $\pm$ 10 (40/72) & 61 $\pm$ 16 (17/28) & 57 $\pm$ 8 (57/100) \\
 & N & N  & 54 $\pm$ 10 (39/72) & 64 $\pm$ 16 (18/28) & 57 $\pm$ 8 (57/100) \\
 \midrule
  \midrule
 BERT-based state of the art~\citep{holzenberger20factoring}  & Y &  N  & 56 $\pm$ 10 (40/72) & 68 $\pm$ 15 (19/28) & 59 $\pm$ 8 (59/100) \\
 Majority baseline & N &  N  & 50 $\pm$ 10 (36/72) & 50 $\pm$ 16 (14/28) & 50 $\pm$ 8 (50/100) \\
 \bottomrule
\end{tabular}

\vspace{1ex}
\caption{GPT-3 performance on SARA with various setups. Below the double line are non-GPT-3 approaches. We report both the 90\% confidence interval and the raw number of accurate answers.}
\label{tab:table_SARA}
\end{center}
\end{table*}

\subsection{Three Approaches}

We try three approaches to prompting GPT-3 to answer the SARA questions.  

First, we use \textbf{4-shot dynamic} prompting.  For each test case, we picked the 4 most similar training cases%, as determined by SARA's name prefixes, which were designed to reflect content
. For these 4, we gave the full text of the case, labeling the "Premise", "Hypothesis", and "Answer", the latter being either "Entailment" or "Contradiction".  Because SARA is balanced, 2 of these 4 training cases had "Answer: Entailment" and the other 2 had "Answer: Contradiction".  Then we appended the "Premise" and "Hypothesis" of the test case, followed by the word "Answer". This is illustrated in the left column of Figure~\ref{fig:prompting-GPT-3-SARA}. 

Second, we use \textbf{zero-shot} prompting.  There we use no training cases at all.  We just put the test case's "Premise" and "Hypothesis" and then "Answer".  This is illustrated in the right column of Figure~\ref{fig:prompting-GPT-3-SARA}. 

Third, we do \textbf{10-shot chain-of-thought} prompting, using a prompt involving 10 of SARA's training examples.  For each of the 10, we have the "Premise",  the "Hypothesis", and "Answer", followed by a human-written chain of thought explaining the relevant reasoning and concluding with the appropriate word "Entailment" or "Contradiction".  The same 10-example prompt is used for all test cases.  This is illustrated in the center column of Figure~\ref{fig:prompting-GPT-3-SARA}. 

GPT-3 does not always complete its answers with "Entailment" or "Contradiction".  In such cases, following \citet{kojima22large}, we then pass to GPT-3 our original prompt, plus GPT-3's response, plus an Answer Extraction prompt reading "Therefore, the answer (Entailment or Contradiction) is".  

\subsection{"Let's think step by step."}

\citet{kojima22large} find substantial improvement on a range of tasks by simply adding "Let's think step by step." at the end of the prompt passed to GPT-3.  They find this forces GPT-3 to explain its reasoning step by step and thus arrive at better answers.  We try each of the zero-shot and 4-shot dynamic prompting experiments both with and without "Let's think step by step."  A column in Table~\ref{tab:table_SARA} indicates whether each row's result used this approach.  Note that we do not use this approach for the 10-shot hand-crafted chain-of-thought, since the 10 examples already contain detailed step-by-step reasoning.  

\subsection{With or Without Statute}

We hypothesize that GPT-3 may have seen the U.S. tax code in its training data, since U.S. federal statutes cannot be copyrighted and thus appear on multiple websites.  So for each experimental setup, we try all the SARA prompts both with and without the text of the statute(s) required to solve all the hypotheses presented, as specified in the "Include Statute" column in Table~\ref{tab:table_SARA}.
In 4-shot dynamic prompting and zero-shot prompting, we use the most relevant section from the statutes. For 10-shot chain-of-thought prompting, we use a set of sections most relevant to the entire test set: 2(b), 63(c)(1) and (6), 152, 3306(a), and 7703.

When we do not include the statutes, we replace all instances of "section" in the SARA premises and hypotheses with "I.R.C. section".  (The U.S. tax code is formally known as the Internal Revenue Code, and "I.R.C." is the standard citation.)  

\subsection{Discussion}

Results for the settings described above are reported in Table~\ref{tab:table_SARA}. As expected, GPT-3 is quite sensitive to the prompt setting, with large variations in performance across settings. The prompt "Let's think step by step." occasionally improves statutory reasoning performance, though not systematically. The BERT-based state of the art we report uses some amount of human-provided structure. Some of our prompt settings surpass that previous state of the art, with far less human input, on the full set of 100 entailment cases.

As measured by Welch's unequal variances $t$-test, our best GPT-3-based model performs significantly better than the BERT-based state of the art ($p \approx 0.038$), and than the majority baseline ($p \approx 0.0011$). With $p<0.05$, only our best model has greater accuracy than the previous state of the art, and only our top two have greater accuracy than the majority baseline.

We note that on nine of our ten experimental setups, GPT-3 performs at least as well on the 28 cases involving no numbers than on the 72 cases involving numbers.  This is expected, since most of the numbers are quite large (e.g. "In 2016, Alice's income was \$310192.") and GPT-3 has a known limitation in doing calculations with large numbers. %\hl{citation that shows this known limitation?}

As expected, GPT-3 performs at chance (0.50) or worse in the zero-shot tests where there is no statute included.  But we were quite surprised to examine GPT-3's step by step reasoning, which shows some (flawed) knowledge of the U.S. tax code.  Here is one example, with our full prompt in regular font, and GPT-3's full response \hlc[GPThighlight]{highlighted}:

\AtBeginEnvironment{quote}{\singlespacing}

\begin{quote}
 Premise: Bob is Alice's brother since April 15th, 2014.

Hypothesis: Alice bears a relationship to Bob under I.R.C. section 152(d)(2)(A).

Answer: Let's think step by step.  \hlc[GPThighlight]{First, I.R.C. section 152(d)(2)(A) states that a taxpayer is allowed to claim a dependent if the dependent is a "brother or sister by whole or half blood." Since Bob is Alice's brother, this would qualify him as a dependent under I.R.C. section 152(d)(2)(A). Therefore, the hypothesis is correct.}
\end{quote}

\vspace{1ex}

Recall that no statutory text was provided in the prompt, so we see that GPT-3 thinks it knows about the U.S. tax code.  But its knowledge is incorrect, since section 152(d)(2)(A) actually states that an individual is a qualifying relative if they are "(A) A child or a descendant of a child" of the taxpayer.  The closest match is in the following subparagraph, section 152(d)(2)(\textbf{B}), which lists "(B) A brother, sister, stepbrother, or stepsister."  In other words, GPT-3 went awry in two ways:  first, it mistakenly referred to the wrong subparagraph; and second, it confused half-siblings with step-siblings.  As a result of this and similar results, we run experiments, described below in Section~\ref{sec:knowsIRC}, to determine how much of the U.S. Code GPT-3 knows.

%\begin{quote}
\begin{figure}[h!]
    \hrule
    \flushleft
        \vspace{1ex}
\textbf{...}

§152. Dependent defined

\textbf{...}

\hspace{1ex} (d) Qualifying relative

\textbf{...}
 
 \hspace{2ex} (2) Relationship
  For purposes of paragraph (1)(A), an individual bears a relationship to the taxpayer described in this paragraph if the individual is any of the following with respect to the taxpayer:
  
 \hspace{3ex}   (A) A child or a descendant of a child.
 
 \hspace{3ex}   (B) A brother, sister, stepbrother, or stepsister.
 
 \hspace{3ex}   (C) The father or mother, or an ancestor of either.
 
 \hspace{3ex}   (D) A stepfather or stepmother.

\textbf{...}

Premise: Charlie is Bob's father since April 15th, 1995. Dorothy is Bob's mother. Alice married Charlie on August 8th, 2018.

Hypothesis: Alice bears a relationship to Bob under section 152(d)(2)(C).

Answer:\hlc[GPThighlight]{  In 2018, Alice is the wife of Charlie. Charlie is Bob's father. Therefore, Alice is Bob's stepmother. Therefore, the answer is Entailment.}
%\end{quote}
\hrule
\caption{Example of SARA getting the wrong answer, because it confused section 152(d)(2)(C) with (d)(2)(D).  This example came from the 10-shot chain-of-thought prompting with the statutes included.  We elided much of the prompt for brevity.}
% We refer the task details to \cref{sec::tasks}.}

    \label{fig:sara_example}
\end{figure}

GPT-3's tendency to mistakenly refer to the wrong part of a statute is not limited to prompts that do not include the text of the statutes.  It happens even when we include the statute in the prompt itself. An example is shown in Figure~\ref{fig:sara_example}.  Given the facts in the prompt, GPT-3 correctly reasons that Alice is Bob's stepmother.  But then it confuses section 152(b)(2)(D) with (b)(2)(\textbf{C}) and gives the wrong answer.  Because of errors like these, we set out in Section~\ref{sec:synthetic} to test GPT-3's ability to reason on synthetic statutes that it has never seen before.  

% \textbf{Include error analysis.  Specifically discuss the strong bias towards finding Entailment.}

\subsection{Does GPT-3 know about SARA?}

SARA was first made available on the internet in 2020, so it is possible that SARA was in GPT-3's training data.  This, in turn, might have biased the results of these experiments.  To test this possibility, we first asked GPT-3 "What is the StAtutory Reasoning Assessment (SARA)?" and it gave an answer that it was a multiple choice test for "law students, lawyers, and other legal professionals." (We know of no such test.)  We then tried passing 20 randomly selected SARA cases followed by the text "Where does the text above come from?"  None of the answers remotely implicated SARA.  The modal answer (5 of 20) was that the language was from the IRS website.  

\section{GPT-3's awareness of the U.S. Code}
\label{sec:knowsIRC}

As discussed earlier, GPT-3's responses to zero-shot SARA prompts without any statutory text demonstrate that it has some, possibly imperfect, knowledge of the U.S. tax code.  In this section we probe what GPT-3 knows about the tax code and U.S. Code in general.  The U.S. Code or U.S.C. is the official compilation of permanent U.S. federal statutes.  It is organized into 53 numbered titles, each corresponding to a particular topic.  Title 26, also called the Internal Revenue Code, is the U.S. tax code.  

We analyze GPT-3's knowledge of the U.S. Code in two directions.  One is to give GPT-3 the full text of a section of the U.S. Code and ask it to identify its origin, including title and section.  The other is to ask GPT-3 to recite the text of a particular section and compare its answer to the actual text.

\subsection{Given statutory text from the U.S. Code, GPT-3 can often identify the title number}

For each of the 53 titles of the U.S. Code, we randomly selected 10 sections whose bodies are between 100 and 1000 words.  The lower limit of 100 words was set to exclude statutes with little semantic content, such as 26 U.S.C. \S 8003, which in its entirety reads "The Joint Committee shall elect a chairman and vice chairman from among its members."  The upper limit of 1000 words was set to avoid running into GPT-3's hard limit of 4000 total tokens, with each word potentially consisting of multiple tokens.  

For each section, we passed a prompt consisting of the text of the statute (\textbf{not} including the section's name, number, or title), followed by a newline and the text "Where is the text above from?".  Once we had a response, following \citet{kojima22large}, we passed the original prompt and response, plus the text "So is it from the U.S. Code? The answer (Yes or No) is".  If the answer was Yes, we passed all the previous prompts and responses concatenated, plus the text "What title of the U.S. Code is it from? The answer (arabic numerals) is".  If we got the correct answer, we passed all the previous prompts and responses concatenated, plus the text "What section of title \textit{N} of the U.S. Code is it from? The answer (arabic numerals) is" where \textit{N} is the proper title.  The results from our 530 tests are in Table~\ref{tab:IDing_USC}. 

\begin{table}[b]

\begin{center}
\begin{tabular}{ rl } 
 \toprule
 Not from U.S. Code & 12\% (64) \\ 
 Wrong title & 25\% (133)  \\ 
 Right title, wrong section & 50\% (265) \\ 
 Title and section correct & 13\% (68)  \\ 
 \bottomrule
\end{tabular}
\end{center}
\vspace{1ex}

\caption{GPT-3's answers when given the text of the body (but not the name, section number, or title number) of a section from the U.S. Code.}
\label{tab:IDing_USC}
\end{table}

Of the 265 sections for which GPT-3 correctly identified the title but could not correctly identify the section, it identified a numerical section for 250 of the 265.  Of these 250, for 36 the section was off by just one (e.g. GPT-3 identified text as from \S 104 when it was actually from \S 103).  For exactly 100 sections, GPT-3 was off by nine or less (e.g. GPT-3 identified text as from \S 161 when it was actually from \S 170).  

Even many of the 64 prompts to which GPT-3 responded "No" when asked "So is it from the U.S. Code? The answer (Yes or No) is" are understandable.  For example, GPT-3 said that the text of 34 U.S.C. \S10707 is not from the U.S. Code.  Rather, GPT-3 responded, "The text above is from the Comprehensive Opioid Abuse Grant Program Evaluation Act of 2016."  The section actually was added by the Comprehensive Addiction and Recovery Act of 2016.  GPT-3 got the title of the Act slightly wrong and entirely missed its subsequent codification into the U.S. Code.  

Of particular relevance to understanding GPT-3's performance on SARA, we separately asked GPT-3 to identify the text of the nine curated statutes in SARA.  GPT-3 correctly identified the title and exact section for eight of the nine.  (It incorrectly identified 26 U.S.C. \S3301 as \S3111.)  This high performance makes sense, given that the sections used in SARA are central to individuals' tax calculations and thus more likely to be talked about extensively in the training corpus for GPT-3.  

\subsection{Given a U.S. Code citation, GPT-3 can recite plausible but incorrect statutory language}

We just saw that GPT-3, when given statutory text from the U.S. Code, is decent at identifying it as such (89\% correct) and even identifying its title number (63\% correct).  We now turn to the flip side: given a citation from the U.S. Code, can GPT-3 recite the text?

For 10 randomly selected statutes in each of the 53 titles of the U.S. Code, we prompted GPT-3 with "The text of \_\_ U.S. Code section \_\_ is:", with the title number in place of the first blank and the section number in place of the second. \footnote{Recall that throughout our experiments we use temperature=0 and top\_P=1.0, which ensures minimal creativity.} 

We never saw GPT-3 exactly recite statutory text.  It always provides an answer in the same style as the U.S. Code, using Congress' formal, dry style and often organizing into numbered subsections, paragraphs, subparagraphs, etc.  Sometimes GPT-3's recitation of a section gets the gist of the section correct, but omits some statutory language and inserts some nonexistent statutory language.  But many times GPT-3 recites text that gives the gist of an entirely incorrect section.  

We aimed to quantify how closely GPT-3 gets at least the gist of sections.  One natural possibility is BLEU, a score used to evaluate machine translations against reference human-generated high-quality transactions \citep{post-2018-call}. But standard BLEU strongly penalizes translations that are shorter than the human-generated reference task, and GPT-3 often recites only the key portions of a statute.   So, we use the metric of \textbf{unpenalized-BLEU}, which is the standard BLEU score without the brevity penalty. Thus, like BLEU, unpenalized-BLEU ranges from 0 (worst) to 100 (best).

Across the 530 sections we asked GPT-3 to recite, we got a mean unpenalized-BLEU score of 7.11 and median of 4.52.  Only 28 of the 530 have an unpenalized-BLEU above 20. For machine translation standards, these are fairly poor results.

Given that GPT-3 often recites roughly correct statutory language for a section different than the one requested, we calculated the unpenalized-BLEU score of the recited text against all the sections in the correct title.  For example, for 17 U.S.C. \S101 we computed the unpenalized-BLEU score of GPT-3's recitation of the section against all 147 sections contained in title 17.  We can then rank the actual U.S.C. sections against GPT-3's prediction. With this, we can measure the rank of the expected section, and thus recall@k. 
Recall@1 is only 1.5\% (8 of the 530 sections sampled) and recall@5 is only 4.7\% (25/530).  

These results led us to evaluate whether GPT-3's recited text reflected the section number at all.  For each of the 530 samples, we normalized the rank of the correct answer within the title by subtracting 1 from the rank and dividing by the number of sections in the title, minus 1. If the correct answer was ranked first for a given section, its normalized recall would thus be 0.0; if GPT-3 is paying no attention whatsoever to the requested section number, we would expect the normalized rank to average 0.5. We observed a mean of 0.48 and median of 0.46, which suggests that GPT-3 pays little attention to the requested section number.   

% - prompt it for relevant SARA statutes (the equivalent in the Tax Code)

% This would have been nice to have, but there isn't time: - Time permitting, I could identify the year with the highest BLEU score.  

\section{GPT-3 struggles with simple reasoning on synthetic statutes}
\label{sec:synthetic}

We know of no technique to determine how much of GPT-3's flawed knowledge of the U.S. Code impacts its performance on SARA, which is based on nine sections drawn from the U.S. Code.  To test GPT-3's ability to reason on statutes it has definitely never seen before, we wrote code to create entirely synthetic statutes, using the same numbering style (i.e. subsection, paragraph, etc.) as the U.S. Code.  We then systematically prompt GPT-3 with the basic statutory reasoning task of giving it one fact and asking it to determine whether a particular subsection applies.\footnote{In this section, as throughout this paper, all our calls to GPT-3 have temperature at 0.0 and the top\_P at 1.0 to maximize reproducibility and minimize hallucination and wandering off topic. }  These statutory-reasoning prompts are similar to SARA, involving self-contained statutes, facts, then simple questions. Also, we take inspiration from the setting of~\citet{saparov22language}.

\subsection{Creation of Synthetic Statutes}

An example of one of our synthetic statutes appeared in Figure \ref{fig:simple_gpt3_fail}.  To avoid any ambiguity, the statutes involve repeated application of the most basic logical form $A \implies B$, such as \textit{parkinse} $\implies$ \textit{rolang}.  In other words, the statutes define terms (rather than, say, imposing duties).  Many real-world statutes consist of precisely this sort of term defining, such as 26 U.S.C. \S 152(a)(1) which provides \textit{qualifying child} $\implies$ \textit{dependent}. 

The statutes are defined in terms of \textit{depth} and \textit{width}.  The width is the number of times, for each term $B$, a relation of $A \implies B$ is stated.  The depth is the maximum number of times this logical form can be applied.  For example, the synthetic statutes in Figures \ref{fig:simple_gpt3_fail} and \ref{fig:statute_fail_prose_works} are both 2-wide and 2-deep.  Meanwhile, the synthetic statute in Figure~\ref{fig:twoshotfail} is 2-wide and 3-deep.  Internally, our code represents these statutes using trees.  

Our synthetic statutes are much simpler than many sections of the U.S. Code.  They involve no negation or ambiguity, and they are entirely balanced, with all leaf nodes being the same depth from the root, and all non-leaf nodes having the same number of children.  In contrast, many sections of the U.S. Code have depth 5, with subsections, paragraphs, subparagraphs, clauses, and subclauses, such as 26 U.S.C. \S2.

\begin{figure}
 % \small
    \hrule
    \flushleft
        \vspace{1ex}

    Section 1001.  Definition of Infarber.

    (a) General rule

    The term "infarber" means-

    \hspace{4ex} (1) any purentiable, or

    \hspace{4ex} (2) any digirderasters.

    (b) Purentiable

    The term "purentiable" means-

    \hspace{4ex} (1) any packle, or

    \hspace{4ex} (2) any amperseced.

    (c) Digirderasters

    The term "digirderasters" means-

    \hspace{4ex} (1) any artion, or

    \hspace{4ex} (2) any irtityrating.
      
    \vspace{1ex}

    Emma is an artion. Is section 1001(b) applicable to Emma? Let's think step by step.

    \vspace{1ex}

    \hlc[GPThighlight]{Section 1001 defines the term "infarber" as either a "purentiable" or a "digirderasters". Emma is an artion, which is a type of digirderasters. Therefore, section 1001(b) is applicable to Emma.}

    \vspace{1ex}

    Therefore, the answer (Yes or No) is \hlc[GPThighlight]{Yes.}

    \vspace{1ex}

    \hrule

    \flushleft
        \vspace{1ex}
        
Sentence 1: The term "infarber" means any purentiable or any digirderasters.
    
    \vspace{1ex}
    
Sentence 2: The term "purentiable" means any packle or any amperseced.
    
    \vspace{1ex}

Sentence 3: The term "digirderasters" means any artion or any irtityrating.

    \vspace{1ex}

Emma is an artion. Is sentence 2 applicable to Emma? Let's think step by step.

    \vspace{1ex}

\hlc[GPThighlight]{Sentence 2 states that the term "purentiable" means any packle or any amperseced. Emma is an artion, which is a type of digirderaster. Therefore, sentence 2 is not applicable to Emma.}

    \vspace{1ex}

Therefore, the answer (Yes or No) is\hlc[GPThighlight]{ No.}
    \vspace{1ex}
    \hrule

\caption{Top: an example of GPT-3 incorrectly answering a zero-shot question about a 2-wide, 2-deep statute.  Bottom: GPT-3 correctly answering the semantically-identical question where the synthetic statute was written in sentence form. Text generated by GPT-3 is \hlc[GPThighlight]{highlighted}. Note that sentence 1 corresponds exactly to section 1001(a), sentence 2 to 1001(b), and sentence 3 to 1001(c).}
% We refer the task details to \cref{sec::tasks}.}

    \label{fig:statute_fail_prose_works}

\end{figure}

 To ensure we are testing GPT-3 on synthetic statutes it has definitely never seen before, we take two approaches to generating the terms being defined.  The first is to use English \textbf{nonces} generated by a nonce generator; examples include roland, parkinse, and oxideney.  The second is to use random \textbf{ids} consisting of one letter plus the same numeral repeated twice; examples include s88, f77, m55, and a22.  

\subsection{Zero-Shot Experimental Setup}

For zero-shot prompting involving our synthetic statutes, as shown in Figure~\ref{fig:simple_gpt3_fail}, the prompt we give to GPT-3 consists of four concatenated strings:

\begin{table*}
\begin{tabular}{ l l l } 
 \toprule
 {\bf phrasing} & {\bf example} &{\bf accuracy (\%)} \\ 
 \midrule
Is  \textit{S} applicable to \textit{N}? & Is section 1001(b) applicable to Alexis? & 77 (2303/3000)  \\ 
 \midrule
Does  \textit{S} apply to \textit{N}? & Does section 1001(b) apply to Alexis? & 74 (2220/3000)  \\ 
 \midrule
Does \textit{S} apply to \textit{N}, making her/him a \textit{T}? & Does section 1001(b) apply to Alexis, making her a rolang? & 54 (540/1000)  \\ 
 \midrule
Does \textit{S} apply to make \textit{N} a \textit{T}?  & Does section 1001(b) apply to make Alexis a rolang? & 58 (576/1000)  \\ 
 \midrule
Is \textit{N} a \textit{T} because of \textit{S}? & Is Alexis a rolang because of section 1001(b)? &  57 (565/1000) \\
 \midrule
Is \textit{N} a \textit{T} owing to \textit{S}? & Is Alexis a rolang owing to section 1001(b)? & 54 (535/1000) \\
 \midrule
Is \textit{N} a \textit{T} as per \textit{S}? & Is Alexis a rolang as per section 1001(b)? & 52 (519/1000) \\ 
 \bottomrule
\end{tabular}

\vspace{1ex}

\caption{Results from different phrasing styles of the same question, all measured on nonce 2-wide, 2-deep synthetic statutes.  The person is \textit{N}, and \textit{S} is the  subsection.  \textit{T} is the top-level term being defined in the synthetic statute.  Note that in all the synthetic prompts we generate, person \textit{N} is a \textit{T}; the only question is whether that happens through \textit{S}.  Note that questions including \textit{T} uniformly perform worse than the two without.  We ran an extra 2000 experiments to choose between the top two phrasings.}
\label{table:phrasing_synthetic}
\end{table*}

\begin{itemize}
   \item the synthetic statute
   \item a single fact in the form "\_\_ is a \_\_.", as with "Alexis is a portle."  The name is randomly chosen from a set of 30 names (15 female, 15 male).\footnote{https://www.ssa.gov/oact/babynames/decades/names2000s.html} The second part of the fact (e.g. portle) is always a term present somewhere in the synthetic statute.     
   \item a question about whether a subsection applies (e.g. "Is section 1001(c) applicable to Alexis?").  This subsection is randomly chosen, although it is never a leaf, meaning it is never a subsection taking up only a single line.  We chose the phrasing "Is \_\_ applicable to \_\_?" after trying a variety of different ways of phrasing the question, with the results shown in Table~\ref{table:phrasing_synthetic}.   
   \item the text "Let's think step by step.", where \citet{kojima22large} found that adding this at the end of the prompt maximizes performance on a variety of reasoning tasks. 
\end{itemize}

We prompt GPT-3 with this string to get an initial response. To derive a definitive answer we then call GPT-3 a second time with the prompt being the original prompt, plus the initial response, and the text "Therefore, the answer (Yes or No) is".  This follows \citet{kojima22large} and is often necessary to force GPT-3 to give a clear answer.  Examples of this prompting can be seen in Figure \ref{fig:statute_fail_prose_works}.  

\begin{table}[b]
\begin{tabular}{ l c c l l } 
 \toprule
 & & & {\bf accuracy on} & {\bf accuracy on} \\ 
 {\bf term} & {\bf width} & {\bf depth} & {\bf statutes (\%)} & {\bf sentences (\%)} \\ 
 \midrule
 nonce & 2 & 2 & 78 (779/1000) &  79 (793/1000) \\ 
 ids & 2 & 2 & 78 (778/1000) & 83 (834/1000) \\ 
\midrule
nonce & 3 & 2 &  70 (698/1000) & 72 (718/1000) \\ 
 ids & 3 & 2 & 68 (682/1000) & 69 (692/1000) \\ 
 \midrule
nonce & 4 & 2 & 69 (688/1000) & 72 (723/1000)  \\ 
 ids & 4 & 2 & 66 (663/1000) & 69 (691/1000) \\ 
 \midrule
nonce & 2 & 3 & 75 (754/1000) & 75 (747/1000)  \\ 
 ids & 2 & 3 & 77 (771/1000) & 66 (662/1000) \\ 
 \midrule
 nonce & 3 & 3 & 75 (374/500) & 74 (371/500) \\ 
 ids & 3 & 3 & 70 (350/500) & 66 (331/500) \\ 
 \bottomrule

\end{tabular}
\vspace{1ex}

\caption{Zero-shot accuracy of GPT-3 on answering whether a  statutory section applies to Alice, given a single fact about Alice. The column "term" indicates whether the statutory terms were nonces like "ansgivath" and "propial" or ids like "f55" and "q11". The rightmost column is GPT-3's accuracy answering questions based on sentences that are semantically-identical to the synthetic statute, as seen in Figure \ref{fig:statute_fail_prose_works}.  }
\label{table:synthetic_0shot}
\end{table}

To test whether GPT-3 has a problem specifically with reasoning over statutes, our code also generates semantically-identical sentence versions of each synthetic statute.  For each defined term in the synthetic statute, our code creates one sentence with the exact same definitional language as the corresponding part of the statute.  These sentences are numbered as "Sentence 1", "Sentence 2", etc. for reference in determining whether a particular sentence is applicable.  For each zero-shot problem posed to GPT-3, we test the semantically-identical sentence version of the same statute and question.  An example appears at the bottom of Figure \ref{fig:statute_fail_prose_works}.  For these sentence versions, the prompt again consists of four parts:

\begin{itemize}
    \item the numbered sentences expressing the same definitions as the synthetic statute.
    \item the same fact as in the statutory version (e.g. "Alexis is a portle.")
    \item a question about whether a sentence applies (e.g. "Does sentence 3 apply to Alexis?").  
   \item the text "Let's think step by step." 
\end{itemize}

We run all tests using both nonce and ids terms. We also run all tests using both the statute version of the definitions and the semantically-identical sentence-based versions.  We balance all tests with equal numbers having positive groundtruth (e.g. section 1001(b) is not applicable) and negative groundtruth (e.g. section 1001(b) is applicable).  

The results are in Table \ref{table:synthetic_0shot}.  Our primary finding is that GPT-3 performs poorly, with performance around 78\% for even the simplest statutes, like that in Figure~\ref{fig:simple_gpt3_fail}, that are 2-wide, 2-deep.  Performance declines further as the statutes get wider or deeper.  

We notice comparable performance between statutes with terms that are nonces like \emph{portle} and statutes with terms that are ids like \emph{m77}.  This makes sense, since both types of synthetic statutes were designed to be entirely novel to GPT-3.  

We also observe comparable performance whether the statute is presented in statute version or semantically-identical sentence-based versions.  GPT-3 correctly answers 6537 of the 9000 statute questions but 6562 of the 9000 sentence-based versions.  

GPT-3's errors are overwhelmingly false positives, meaning it concludes that a section or sentence applies when it actually does not.  In the 9000 nonce runs, there were 2272 errors.  Of these, 2204 were false positives and 61 false negatives.  (There were also 7 where GPT-3 did not give a definitive answer.)  

\subsection{Two-Shot Experimental Setup}

\begin{figure}[b]
 % \small
    \hrule
    \flushleft
        \vspace{1ex}

Section 1001.  Definition of Bowlery.

(a) General rule

The term "bowlery" means-

\hspace{2ex}  (1) any waitormenteed, or

\hspace{2ex}  (2) any kiterrupider.  

(b) Waitormenteed

\hspace{2ex}  (1) General rule

 \hspace{2ex} The term "waitormenteed" means-

   \hspace{4ex}    (A) any redeba, or

   \hspace{4ex}   (B) any dischieviders.

\hspace{2ex}   (2) Redeba

 \hspace{2ex} The term "redeba" means-

  \hspace{4ex}   (A) any ersubs, or

 \hspace{4ex}    (B) any pushotyptopses.

\hspace{2ex}   (3) Dischieviders

 \hspace{2ex}  The term "dischieviders" means-

 \hspace{4ex}    (A) any nookede, or
 
 \hspace{4ex}    (B) any chastiles.

(c) Kiterrupider

\hspace{2ex}  (1) General rule

\hspace{2ex}   The term "kiterrupider" means-

 \hspace{4ex}   (A) any bruselers, or

 \hspace{4ex}    (B) any fashiple.

\hspace{2ex}  (2) Bruselers

\hspace{2ex}   The term "bruselers" means-

 \hspace{4ex}   (A) any legimetar, or

 \hspace{4ex}   (B) any exematess.

\hspace{2ex}  (3) Fashiple

\hspace{2ex}  The term "fashiple" means-

\hspace{4ex}    (A) any tanded, or

\hspace{4ex}    (B) any goghts.

    \vspace{1ex}

Hannah is a chastiles. Is section 1001(c)(3) applicable to Hannah? Section 1001(c)(3) says that fashiple means any tanded or any goghts. Hannah is none of these, so section 1001(c)(3) does NOT apply to Hannah.
    \vspace{1ex}

Alyssa is a goghts. Is section 1001(c)(3) applicable to Alyssa? Section 1001(c)(3) says that fashiple means any tanded or any goghts. Alyssa is a goghts, so section 1001(c)(3) does apply to Alyssa.
    \vspace{1ex}

Nicholas is a pushotyptopses. Is section 1001(c)(2) applicable to Nicholas? \hlc[GPThighlight]{Section 1001(c)(2) says that redeba means any ersubs or any pushotyptopses. Nicholas is a pushotyptopses, so section 1001(c)(2) does apply to Nicholas.}
            \vspace{1ex}

    \hrule

\caption{Two-shot prompt with GPT-3 giving an incorrect answer.  This is a 2-wide, 3-deep statute.  Note that the two examples given to GPT-3, involving Hannah and Alyssa, are correct. GPT-3's answer shows is has incorrectly looked at section 1001(b)(2), not (c)(2).}
% We refer the task details to \cref{sec::tasks}.}

    \label{fig:twoshotfail}

\end{figure}

Given GPT-3's poor performance on the zero-shot statutory reasoning discussed above, we turned to whether GPT-3 performs better with two-shot reasoning.  In two-shot reasoning we give two correct examples (one answered Yes, the other answered No), before posing the actual question. All two-shot experiments were run with synthetic statutes where the terms were nonces. (We used neither semantically-identical sentences, nor ids like g11.)  A two-shot example involving a 2-wide, 3-deep statute appears in Figure \ref{fig:twoshotfail}.  

\begin{table}
\begin{tabular}{ c c l l } 
 \toprule
 &  & {\bf two-shot } & {\bf zero-shot} \\  
 {\bf width} & {\bf depth} & {\bf accuracy (\%)} & {\bf accuracy (\%)}\\ 
 \midrule
 2 & 2 & 100 (1000/1000)  & 78 \\ 
 \midrule
 3 & 2 & 98 (982/1000) & 70 \\ 
 \midrule
 4 & 2 & 97 (972/1000) & 69 \\ 
 \midrule
 2 & 3 & 87 (874/1000) & 75 \\ 
 \midrule
 3 & 3 & 81 (405/500) & 75 \\ 

 \bottomrule
\end{tabular}

\vspace{1ex}

\caption{Two-shot accuracy of GPT-3, when the prompt has two example questions with reasoning and correct answers, as in Figure~\ref{fig:twoshotfail}.  The leftmost column reproduces the comparable zero-shot accuracy from Table~\ref{table:synthetic_0shot}. }
\label{table:synthetic_2shot}
\end{table}

Both the correctly answered questions are of the form "\_\_ is a \_\_.  Does section \_\_ apply to \_\_?", followed by a correct explanation and answer.  The Yes-answered question comes first with 50\% probability.  Both questions relate to the same section, which never overlaps with the section asked about in the third question.  All three questions involve different randomly-selected names (e.g. Hannah, Alyssa, Nicholas) and randomly-selected terms (e.g. chastiles, goghts, pushotyptopses).  

We see uniformly better performance with the two-shot model than with the zero-shot model (Table \ref{table:synthetic_2shot}).  With two examples, GPT-3 seems capable of handling 100\% of 2-wide, 2-deep statutes, such as those shown on our first page in Figure~\ref{fig:simple_gpt3_fail}.  As in the zero-shot setting, GPT-3's accuracy tends to decrease as the statutes' depth and width increase.  GPT-3 still performs quite poorly (81\%) on 3-wide, 3-deep statutes,  which themselves pale in comparison to the complexity of many sections in the U.S. Code. 

\section*{Conclusion}

Being able to identify whether a specific subsection of an unfamiliar statute is applicable to a given set of facts is one of the most basic skills required of a lawyer.  We find that, given a very simple 2-wide, 2-deep synthetic statute paired with a single fact and single question (\emph{zero-shot}), GPT-3 had 78\% accuracy, raising doubts about GPT-3's ability to handle basic legal work.  Providing two examples of correct statutory reasoning (\emph{two-shot reasoning}) improves GPT-3's performance, but it still achieves only (81\%) on 3-wide, 3-deep synthetic statutes, which are far less complex than what is found in the U.S. Code. %This complexity is below what is found in the US Code and is not representative of human lawyer performance. 
This poor performance on synthetic statutory reasoning allows us to understand our results applying GPT-3 to the SARA dataset.  These results, at 71\% accuracy, are better than the previous state of the art, but leave significant room for improvement. We hope this work motivates further research into improving the performance of large language models on statutory reasoning. We also look forward to testing new large language models, like GPT-4, on statutory reasoning. 

\section*{ACKNOWLEDGMENTS}

This work has been supported by the U.S. National Science Foundation under grant No. 2204926. Any opinions, findings, and conclusions or recommendations expressed in this article are those of the authors and do not necessarily reflect the views of the National Science Foundation.

%Being able to identify whether a specific subsection of an unfamiliar statute applies to a given set of facts is one of the most basic skills required of a lawyer. In this paper, we ask whether GPT-3 can emulate that skill. First, we test its ability on an established benchmark for statutory reasoning. Experimenting with multiple settings, we find that GPT-3 performs better than previous state-of-the-art. We identify several types of errors, motivating further experiments. Second, we probe GPT-3's knowledge of U.S. statutes: while GPT-3 knows about statutory language, it only has a vague idea of the content of statutes. Finally, we test GPT-3's statutory reasoning abilities with synthetic statutes. We find that, given a simple synthetic statute and a single fact, GPT-3 currently performs poorly. Providing two examples of correct statutory reasoning (2-shot reasoning) improves GPT-3's performance, but it still achieves only 75\% accuracy on 3-wide, 3-deep synthetic statutes. This remains at odds with lawyers' ability to reason reliably with a novel statute, or a modified version of a familiar statute. 

%Going back to our initial question, it is not clear whether GPT-3 can perform statutory reasoning with the standard methods developed by the NLP community. We hope this work motivates further research into how to leverage large language models to perform statutory reasoning, and reasoning in general. %We see at least three possible directions for future research: prompt engineering, broader pre-training, and controlled decoding.

\bibliographystyle{ACM-Reference-Format}
\bibliography{sample-base}



\end{document}